\documentclass{article}

\PassOptionsToPackage{compress}{natbib}
 \usepackage[preprint]{neurips_2025}

\usepackage[utf8]{inputenc} 
\usepackage[T1]{fontenc}    
\usepackage{hyperref}       
\usepackage{url}            
\usepackage{booktabs}       
\usepackage{amsfonts}       
\usepackage{nicefrac}       
\usepackage{microtype}      
\usepackage{xcolor}         
\usepackage{amsmath}
\usepackage{graphicx}

\title{Thermodynamic Performance Limits for Score-Based Diffusion Models}
\author{%
  Nathan X.~Kodama \\
  Case Western Reserve University\\
  Cleveland, OH 44106 \\
  \texttt{nxk281@case.edu} \\
  \And
  Michael Hinczewski \\
  Case Western Reserve University\\
  Cleveland, OH 44106 \\
  \texttt{mxh605@case.edu} \\
}

\begin{document}

\maketitle

\begin{abstract}
We establish a fundamental connection between score-based diffusion models and non-equilibrium thermodynamics by deriving performance limits based on entropy rates. Our main theoretical contribution is a lower bound on the negative log-likelihood of the data that relates model performance to entropy rates of diffusion processes. We numerically validate this bound on a synthetic dataset and investigate its tightness. By building a bridge to entropy rates---system, intrinsic, and exchange entropy---we provide new insights into the thermodynamic operation of these models, drawing parallels to Maxwell's demon and implications for thermodynamic computing hardware. Our framework connects generative modeling performance to fundamental physical principles through stochastic thermodynamics.
\end{abstract}

\section{Introduction}
Score-based diffusion models have achieved remarkable success in generative modeling by learning to reverse a stochastic diffusion process~\citep{Song2021}. Recent advances have exploited physical connections to optimal transport~\citep{Kwon2022,Lipman2022}, critical damping~\citep{Dockhorn2022}, and heat dissipation~\citep{Rissanen2023} to achieve significant performance gains, while others have connected generative processes to Maxwell's demon~\citep{Premkumar2025} and thermodynamic hardware~\citep{Coles2023}.

Extending on pioneering work connecting deep learning with non-equilibrium thermodynamics \citep{Sohl-Dickstein2015}, recent work has highlighted fundamental connections between these models and stochastic thermodynamics, including speed--accuracy tradeoffs derived from entropy production \citep{Ikeda2025}. Our contribution is complementary: we focus on formalizing the analogy to Maxwell's demon and deriving a thermodynamically motivated lower bound on the negative log-likelihood (NLL).

Prior variational treatments provide an evidence lower bound (ELBO) on the log-likelihood
\citep{Huang2021}, which is equivalently an upper bound on NLL, and some analyses give upper
bounds on $\mathrm{KL}(p_{\mathrm{data}}\|p_{\mathrm{model}})$ \citep{Premkumar2025}, again implying upper bounds
on NLL. In contrast, under a consistent plug-in convention where system entropy rates $\dot{S}_{\boldsymbol{\theta}}(t)$ are computed from the learned score, we derive a thermodynamic lower bound on NLL
\[
\mathrm{NLL} \;\geq\; \frac{S_0 + S_1}{2} - \frac{1}{2} \int_0^1 \dot S_{\boldsymbol \theta}(t)\,dt,
\]
where $S_0$ is the entropy of the data and $S_1$ that of the equilibrium distribution. Note, a trivial bound $\mathrm{NLL}\geq S_0$ follows from $\mathrm{KL}\geq 0$: our result strengthens it via $S_1$ and entropy–rate corrections. Because NLL is a widely reported performance metric for diffusion models, this inequality gives a clear limit on achievable performance: no training or sampling procedure can reduce NLL below this thermodynamically motivated floor, distinguishing our bound from the ELBO- and KL-based bounds.

\section{Background}
\subsection{Score-Based Diffusion Models}
Score-based diffusion models learn to reverse a forward diffusion process.
We consider the forward stochastic process $\mathbf{x}_t \in \mathbb{R}^d$ governed by the Itô SDE:
$$
d \mathbf{x}_t=\mathbf{f}(\mathbf{x}_t, t) d t+\mathbf{G}(\mathbf{x}_t, t) d \mathbf{w}_t, \quad t \in [0,T],
$$
where $\mathbf{f}(\mathbf{x}_t, t): \mathbb{R}^d \times [0,T] \rightarrow \mathbb{R}^d$ is the deterministic drift vector, $\mathbf{G}(\mathbf{x}_t,t): \mathbb{R}^d \times [0,T] \rightarrow \mathbb{R}^{d \times m}$ is the stochastic diffusion matrix, and $\mathbf{w}_t$ is an $m$-dimensional standard Wiener process.
The reverse-time diffusion process $\overline{\mathbf{x}}_t:=\mathbf{x}_{\tau}$ with $\tau=T-t$ can be derived~\citep{Anderson1982,Haussmann1986,Song2021}:
$$
d \overline{\mathbf{x}}_\tau=\left[-\mathbf{f}\left(\overline{\mathbf{x}}_\tau, T-\tau\right)+ 2\mathbf{D}(\overline{\mathbf{x}}_\tau,T-\tau) \nabla_{\overline{\mathbf{x}}_\tau} \log p_{\tau}\left(\overline{\mathbf{x}}_\tau\right)\right] d \tau+\mathbf{G}\left(\overline{\mathbf{x}}_\tau, \tau\right) d \mathbf{w}_\tau
$$
where $\mathbf{D}(\overline{\mathbf{x}}_\tau,T-\tau) = \frac{1}{2}\mathbf{G}\left(\overline{\mathbf{x}}_\tau, T-\tau\right) \mathbf{G}\left(\overline{\mathbf{x}}_\tau, T-\tau\right)^{\top}$ and $\nabla_{\overline{\mathbf{x}}_\tau} \log p_{\tau}\left(\overline{\mathbf{x}}_\tau\right)$ is called the score function of the marginal distribution over $\overline{\mathbf{x}}_\tau$. Score-based diffusion models use a deep neural network to approximate the score function: $\mathbf{s}_{\boldsymbol{\theta}}(\mathbf{x},\tau) \approx \nabla_{\overline{\mathbf{x}}_\tau} \log p_{\tau}\left(\overline{\mathbf{x}}_\tau\right)$.

The reverse-time process can be used as a generative model. In particular, \citep{Song2021} model data $\mathbf{x}$, setting $p\left(\mathbf{x}_0\right)=p_{\mathrm{data }}(\mathbf{x})$. Currently, diffusion models~\citep{Song2021} have drift and diffusion coefficients of the simple form $\mathbf{f}\left(\mathbf{x}_t, t\right)=f(t) \mathbf{x}_t$ and $\mathbf{G}\left(\mathbf{x}_t, t\right)=g(t) \mathbf{I}_d$. Generally, $\mathbf{f}$ and $\mathbf{G}$ are chosen such that the marginal, equilibrium density is approximately normal at time $T$, i.e., $p\left(\mathbf{x}_T\right) \approx \mathcal{N}\left(\mathbf{0}, \mathbf{I}_d\right)$. We can then initialize $\mathbf{x}_0$ based on a sample drawn from a complex data distribution, corresponding to a far-from-equilibrium state. While the state $\mathbf{x}_0$ relaxes towards equilibrium via the forward diffusion, we can learn a model $\mathbf{s}_{\boldsymbol{\theta}}\left(\mathbf{x}_t, t\right)$ for the score $\nabla_{\mathbf{x}_t} \log p_t\left(\mathbf{x}_t\right)$, which can be used for generation via the reverse process. If $f$ and $G$ take the simple form from above, the unweighted denoising score matching~\citep{Vincent2011} objective for this task is:
$$
\min _{\boldsymbol{\theta}} \mathbb{E}_{t \sim \mathcal{U}[0, T]} \mathbb{E}_{\mathbf{x}_0 \sim p\left(\mathbf{x}_0\right)} \mathbb{E}_{\mathbf{x}_t \sim p_t\left(\mathbf{x}_t \mid \mathbf{x}_0\right)}\left[\left\|\mathbf{s}_{\boldsymbol{\theta}}\left(\mathbf{x}_t, t\right)-\nabla_{\mathbf{x}_t} \log p_t\left(\mathbf{x}_t \mid \mathbf{x}_0\right)\right\|_2^2\right]
$$

\subsection{Stochastic Thermodynamics}
In stochastic thermodynamics, entropy production quantifies the irreversibility in non-equilibrium processes. Recent work has applied these principles to diffusion models, showing that entropy production constrains achievable speed and accuracy \citep{Ikeda2025}. For a stochastic process, the system entropy production—the rate of change $\dot S(t)$ of its Gibbs entropy $S(t)$—can be decomposed as \citep{Seifert2012}:
$$
\dot{S}(t)=\dot{S}^i(t)+\dot{S}^e(t)
$$
where intrinsic entropy production $\dot{S}^i(t)$ is always non-negative and measures irreversibility. The remaining term, $\dot{S}^e(t)$, is known as an exchange entropy rate for a system connected to a thermal heat bath, since it is related to rate of heat exchange with the bath. Details on how to compute analogous quantities for score-based diffusion models are provided in Appendix~\ref{app:entropy-rates}.

\section{Main Results}
\subsection{Lower Bound on Negative Log-Likelihood}
For an approximate score function $\mathbf{s}_{\boldsymbol{\theta}}(\mathbf x,T-\tau)$, the negative log-likelihood (NLL) satisfies
\begin{equation}
\boxed{\mathrm{NLL} - S_0 \;\;\geq\;\; \frac{1}{2} \left[S_1 - S_0 - \int_0^1 \dot S_{\boldsymbol{\theta}}(T-\tau)\, d\tau\right],}
\end{equation}
where $S_0$ is the entropy of the data distribution, $S_1$ that of the equilibrium (prior), and
$\dot S_{\boldsymbol{\theta}}$ the entropy rate defined by the learned score function. The trivial bound
$\mathrm{NLL}\geq S_0$ follows directly from $\mathrm{NLL} = S(p_{\mathrm{data}}, p_{\boldsymbol{\theta}}) \geq S(p_{\mathrm{data}}) = S_0$, i.e.\ from the non-negativity of $\mathrm{KL}(p_{\mathrm{data}} \| p_{\boldsymbol{\theta}})$. Equality holds only if $p_{\boldsymbol{\theta}} = p_{\mathrm{data}}$; our result strengthens
it by incorporating $S_1$ and entropy-rate corrections. Details of the derivation appear in
Appendix~\ref{app:lower-bound}. Briefly, the bound comes from the definition of negative log-likelihood in the probability flow ODE framework~\citep{Song2021} associated with the above SDEs, followed by applying polarization and Stein's identities combined with the score-based definition of entropy rates.

\subsection{Connection to Maxwell's Demon and Entropy Rates}
The Maxwell's Demon thought experiment involves an external controller that selectively manipulates systems to lower their entropy. Score-based models operate analogously to Maxwell's Demon: the neural network measures the system state during training (forward process) and uses this information to decrease entropy during generation (reverse process). The reverse process mirrors how Maxwell’s Demon manipulates particles in hot and cold reservoirs to impose order~\citep{Coles2023}.

We consider the special case of drift-less diffusion, $d\mathbf x_t = g(t) d \mathbf w_t$. For a score network that reverses drift-less diffusion, the intrinsic entropy production rate is 
\begin{equation}
\dot{S}^i_{\boldsymbol{\theta}}(T-\tau)=\frac{g(T-\tau)^2}{2} \mathbb{E}\left[\left\|\mathbf s_{\boldsymbol{\theta}}\left(\overline{\mathbf x}_\tau, T-\tau\right)\right\|^2\right].
\end{equation}
While the drift-less forward process has no exchange entropy, the reverse process has exchange-entropy rate that is
$$
\dot{S}^{\mathrm{e}}_{\boldsymbol{\theta}}(T-\tau)=\mathbb{E}\left[\nabla_{\overline{\mathbf x}_\tau} \cdot \tilde {\mathbf f}_{\boldsymbol{\theta}}\left(\overline{\mathbf x}_\tau, T-\tau\right)\right].
$$
For the score network controlled-forward process (see Appendix~\ref{app:controlled-forward}), the drift is  $\tilde {\mathbf f}_{\boldsymbol{\theta}}\left(\overline{\mathbf x}_\tau, T-\tau\right)=g(T-\tau)^2 \mathbf{s}_{\boldsymbol{\theta}}(\overline{\mathbf{x}}_\tau, T-\tau)$, so 
\begin{align*}
    \dot S^e_{\boldsymbol{\theta}}(T-\tau) &= g(T-\tau)^2\mathbb{E}\left[\nabla_{\overline{\mathbf{x}}_{\tau}} \cdot \mathbf{s}_{\boldsymbol{\theta}}(\overline{\mathbf{x}}_\tau, T-\tau)\right] \\&= -g(T-\tau)^2\mathbb{E}\left[||\mathbf{s}_{\boldsymbol{\theta}}(\overline{\mathbf{x}}_\tau, T-\tau)||^2\right] = -2 \dot{S}_{\boldsymbol{\theta}}^{i}(T-\tau),
\end{align*}
where we have used Stein's identity (see Appendix~\ref{app:steins-identity}). Thus, the system entropy rate is
\begin{align*}
\dot{S}_{\boldsymbol{\theta}}(T-\tau)&=\dot S^i_{\boldsymbol{\theta}}(T-\tau)+\dot S^e_{\boldsymbol{\theta}}(T-\tau) \\&=\dot S^i_{\boldsymbol{\theta}}(T-\tau) - 2 \dot S^i_{\boldsymbol{\theta}} (T-\tau) = - \dot S^i_{\boldsymbol{\theta}}(T-\tau),
\end{align*}
which means that a good score network must completely reverse the forward process. This connects the score model directly to thermodynamic entropy rates and the neural network's outputs to Maxwell's Demon.

\section{Numerical Results}
We validate our theoretical predictions using synthetic 8-bit grayscale images with uniformly distributed pixel values between $0$ and $1$. Our numerical experiments use a score-based diffusion model with a U-Net architecture to approximate the score function $\mathbf s_{\boldsymbol{\theta}}(\mathbf x,t)$. We compute exact negative log-likelihood values via the probability ODE framework and measure entropy rates directly from the trained neural network's score function approximation, enabling direct comparison with our theoretical predictions.

\begin{figure}[h!]
  \centering
  \includegraphics[width=\textwidth]{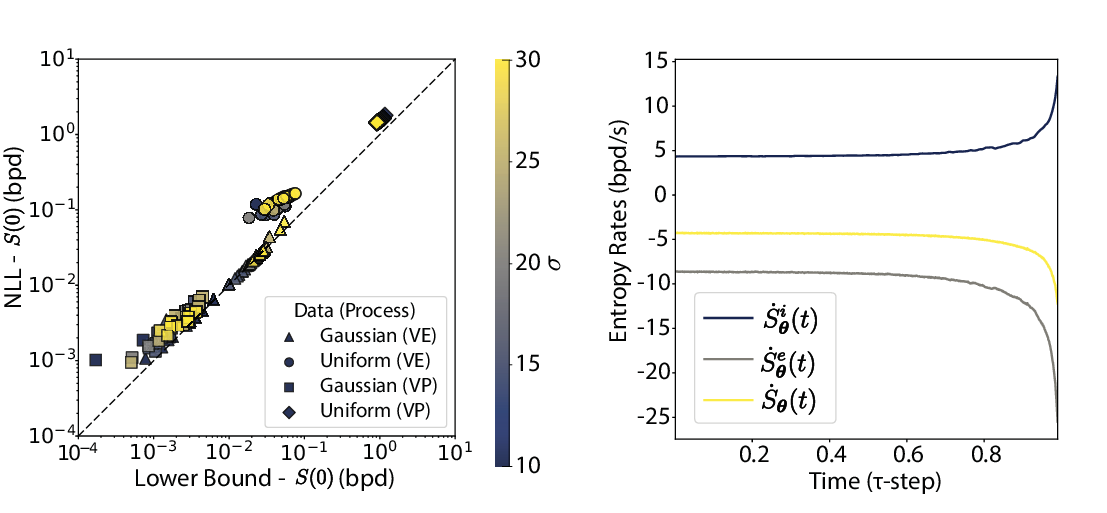}
  \caption{Comparison between the NLL and theoretical lower bound  across diffusion model configurations. (Left) NLL values versus the lower bound in Eq.~(1) (dashed) for Gaussian and Uniform data under both variance exploding (VE) and variance preserving (VP) processes. Marker shape denotes data distribution and process, while color indicates the noise parameter $\sigma \in [10,30]$. (Right) Entropy rates (intrinsic $\dot S^i_{\boldsymbol \theta}$, exchange $\dot S^e_{\boldsymbol \theta}$, and system $\dot S_{\boldsymbol \theta}$) estimated from the score network yielding the best NLL in the Uniform (VE) case, confirming the predicted $2{:}1$ ratio $\dot S^e_{\boldsymbol \theta} = -2 \dot S^i_{\boldsymbol \theta}$.}
  \label{fig:lb}
\end{figure}

In Figure~\ref{fig:lb}, the left panel exposes the relationship between the negative log-likelihood and lower bound across 5 noise parameters, $\sigma \in \{10,15,20,25,30 \}$, and 10 runs per noise parameter. The theoretical bound consistently hold across all parameters and runs, with tighter bounds correlating with better model performance. We observe strong positive correlations between the negative log-likelihood and the performance gap, quantified by the Pearson coefficient ($r=0.694$, $p<0.001$) and Spearman coefficient ($r_s=0.882$, $p<0.001$). The performance gaps correspond to the squared difference term $||\mathbf s_{\boldsymbol{\theta}}-\mathbf s_\mathrm{true}||^2$ in the exact decomposition of the negative log-likelihood, confirming that models with better score approximations achieve both lower negative log-likelihood and tighter bounds.

Entropy rate estimates (intrinsic $\dot S^i_{\boldsymbol{\theta}}$, exchange $\dot S^e_{\boldsymbol{\theta}}$, and system $\dot S_{\boldsymbol{\theta}}$) computed from the score neural network yielding the best performance are presented in the right panel of Figure~\ref{fig:lb}. These empirical measurement validate our theoretical predictions: the intrinsic entropy production rate $\dot S^i_{\boldsymbol{\theta}}(T-\tau)$ remains positive throughout the controlled process, the exchange entropy rate maintains the predicted 2:1 ratio, $\dot S_{\boldsymbol{\theta}}^e(T-\tau) = -2 \dot S^i_{\boldsymbol{\theta}}(T-\tau)$, and the system entropy rate $\dot S_{\boldsymbol{\theta}}(T-\tau) = -\dot S^i_{\boldsymbol{\theta}}(T-\tau)$ confirms that the score network successfully reverses the forward diffusion process by maintaining negative system entropy production.

\section{Conclusion}
Our work establishes fundamental connections between generative modeling and statistical physics, elaborating on pioneering insights connecting deep learning with non-equilibrium thermodynamics~\cite{Sohl-Dickstein2015} and complementing recent analyses of speed--accuracy tradeoffs in diffusion models \citep{Ikeda2025}. Our contributions are to formalize the Maxwell’s demon analogy and derive a lower bound on NLL expressed in terms of entropy rates. Our theoretical framework extends on existing variational bounds~\citep{Huang2021} by deriving a fundamental limit that relates model performance directly to entropy rates in diffusion processes. There are several practical implications and applications.

\textbf{Thermodynamic Computing}. Our results suggest fundamental limits that may be exploited in thermodynamic hardware. In the current formulation, entropy rates are defined via mathematical analogy to thermodynamics. However, when realized on thermodynamic hardware~\citep{Coles2023}, entropy rates become physical quantities and the bound becomes a target, extending connections to Maxwell’s demon~\citep{Premkumar2025} into practical hardware design principles.

\textbf{Performance Analysis}. Entropy rates provide new diagnostics for model behavior, complementing existing metrics with physically motivated quantities that reveal fundamental trade-offs. In particular, the mathematical connection to Maxwell’s Demon in terms of entropy rates not only provides a conceptual framework for understanding the operation of score-based diffusion models, but also enables us to estimate the amount of entropy the score network removes from the system during the reverse process. This perspective clarifies the thermodynamic role of the score network and highlights entropy reduction as a measurable quantity that links model performance to physical limits.

\textbf{Control Generative Models}. Minimizing entropy production while maintaining model quality could lead to faster sampling and training. Connections to optimal transport theory~\citep{Kwon2022,Lipman2022} and thermodynamic uncertainty principles suggest design principles for designing more controllable and efficient diffusion models.

We have established rigorous connections between score-based diffusion models and non-equilibrium thermodynamics, providing theoretical insights and practical tools. Our lower bound based on entropy rates sets fundamental performance limits, while the mathematical description of Maxwell's demon in terms of entropy rates offers a framework for understanding the operation of score-based generative models using tools from stochastic thermodynamics.

\newpage
\bibliographystyle{unsrtnat}
\bibliography{references}{}


\appendix
\newpage
\section{Entropy Rates for Score-Based Diffusion Models}\label{app:entropy-rates}
Seifert’s original formulation \citep{Seifert2012} and subsequent applications to diffusion models \citep{Ikeda2025} motivate the mathematical framework we adopt here.

\subsection{Current-score-drift identity}
For the general overdamped SDE $d \mathbf x_t=\mathbf f\left(\mathbf x_t, t\right) d t+g(t) d \mathbf w_t$, the probability current is
$$
\mathbf J(\mathbf x, t)=\mathbf f(\mathbf x, t) p_t(\mathbf x)-D(t) \nabla_{\mathbf x} p_t(\mathbf x)=p_t(\mathbf x)[\mathbf f(\mathbf x, t)-D(t) \mathbf s(\mathbf x, t)],
$$
where we have used $g(t)^2 = 2 D(t)$, $\mathbf s(\mathbf x)=\nabla \log p_t(\mathbf x)$ and $\nabla p_t(\mathbf{x})=p_t(\mathbf x)\nabla \log p_t(\mathbf x)=p_t(\mathbf x)\mathbf s(\mathbf x)$. The local velocity field is defined as
$$
\mathbf v(\mathbf x, t):=\mathbf f(\mathbf x, t)-D(t) \mathbf s(\mathbf x, t)=\mathbf J(\mathbf x,t)/p_t(\mathbf{x})
$$

\subsection{Intrinsic entropy-production rate}
\cite{Seifert2012}'s original expression for the intrinsic entropy production rate is given by
$$
\dot{S}^i(t)=\int \frac{\left\|\mathbf{J}(\mathbf{x},t)\right\|^2}{D(t) p_t(\mathbf{x})} \mathrm{d} \mathbf{x}=\frac{1}{D(t)}\int p_t(\mathbf x) \left\|\mathbf v(\mathbf x,t)\right\|^2 d \mathbf x.
$$

In expectation notation,
\begin{align*}
\dot{S}^i(t)&=\frac{1}{D(t)} \mathbb{E}\left[\|\mathbf v(\mathbf x,t)\|^2\right]=\frac{2}{g(t)^2} \mathbb{E}\left[\|\mathbf f(\mathbf x,t)-\frac{g(t)^2}{2} \mathbf s (\mathbf x,t)\|^2\right]\\&=\frac{1}{2g(t)^2} \mathbb{E}\left[\|2\mathbf f(\mathbf x,t)-g(t)^2 \mathbf s (\mathbf x,t)\|^2\right]
\end{align*}

\subsection{Exchange (medium) entropy-flow rate} 

Seifert defines the entropy component of the medium surrounding  a system (related to the heat dissipated into that medium) through the work done by the force $F(\mathbf x, t)$ on the system at some time-dependent temperature $T(t)$,
$$
\dot{S}^m(t)=\frac{1}{T(t)} \int \mathbf F(\mathbf x, t) \cdot \mathbf J(\mathbf{x}, t)d\mathbf x=\frac{1}{D(t)} \int  \mathbf f(\mathbf x, t) \cdot \mathbf J(\mathbf x, t) d \mathbf x
$$
In order to make the analogy between the diffusion algorithm and a physical system, we imagine a mobility (inverse friction) constant $\mu$ and corresponding Einstein relation $D(t)=\mu T(t)$, allowing us to write the drift term $\mathbf f=\mu \mathbf F$ in the SDE $d \mathbf x_t = \mu \mathbf F(\mathbf x,t) + g(t) d \mathbf w_t$.

The exchange/flow rate of entropy into the system is just the negative of the one into the medium,
\begin{align*}
\dot{S}^e(t)=-\dot{S}^m(t)&=-\frac{1}{D(t)} \mathbb{E}\left[\mathbf{f}(\mathbf x,t) \cdot (\mathbf f(\mathbf x,t) - D(t)\mathbf s (\mathbf x,t))\right]\\&=-\frac{1}{D(t)} \mathbb{E}\left[\|\mathbf f (\mathbf x,t)\|^2\right]+\mathbb{E}[\mathbf f(\mathbf{x},t) \cdot \mathbf s(\mathbf x,t)] \\&=-\frac{2}{g(t)^2} \mathbb{E}\left[\|\mathbf f (\mathbf x,t)\|^2\right]+\mathbb{E}[\mathbf f(\mathbf{x},t) \cdot \mathbf s(\mathbf x,t)], 
\end{align*}
where in the first line we have use the fact that $\mathbf{J}(\mathbf{x},t) = \mathbf{v}(\mathbf{x},t) p_t(\mathbf{x})$.

\subsection{System entropy rate}
Combining the expressions for $\dot{S}^i$ and $\dot{S}^e$, expanding the square and canceling terms gives the simplified equation for the (total) system entropy rate:
\begin{align*}
\dot{S}(t)&=\dot{S}^i(t)+\dot{S}^e(t)\\&=\frac{1}{D(t)} \mathbb{E}\left[\|\mathbf f(\mathbf x,t)-D(t) \mathbf s (\mathbf x,t)\|^2\right]-\frac{1}{D(t)} \mathbb{E}\left[\|\mathbf f (\mathbf x,t)\|^2\right] +\mathbb{E}[\mathbf f(\mathbf{x},t) \cdot \mathbf s(\mathbf x,t)]\\&=-\mathbb{E}[\mathbf f(\mathbf{x},t) \cdot \mathbf s(\mathbf x,t)]+D(t) \mathbb{E}\left[\|\mathbf s(\mathbf x,t)\|^2\right]\\&=\mathbb{E}[\nabla_\mathbf{x}\cdot \mathbf f(\mathbf{x},t)]+\frac{g(t)^2}{2} \mathbb{E}\left[\|\mathbf s(\mathbf x,t)\|^2\right].
\end{align*}
where we have used Stein's identity for $\mathbb{E}[\nabla_\mathbf{x}\cdot \mathbf f(\mathbf{x},t)]=-\mathbb{E}[\mathbf f(\mathbf{x},t) \cdot \mathbf s(\mathbf x,t)]$ (see Sec.~\ref{app:steins-identity}).

\section{Lower Bound for Negative Log-Likelihood}\label{app:lower-bound}
\subsection{Log-Likelihood from Probability Flow ODE}
For all diffusion processes, there exists a corresponding deterministic process called the probability flow ODE whose trajectories share the same marginal probability densities $\left\{p_t(\mathbf{x})\right\}_{t=0}^T$ as the SDE~\cite{Song2021}. For the case of $d \mathbf{x}_t= \mathbf{f}(\mathbf x_t,t)dt + g(t) \mathrm{d} \mathbf{w}_t$, where $ g(t)= \sigma^t$, the probability flow ODE is
$$
d \mathbf{x}_t=\left[\mathbf{f}(\mathbf x_t,t)-\frac{1}{2} g(t)^2 \nabla_{\mathbf{x}_t} \log p_t(\mathbf{x}_t) \right] dt.
$$

The probability flow ODE has the following form when we approximate the score  with the score neural network model $\mathbf{s}_{\boldsymbol{\theta}}(\mathbf{x}_t, t)\approx\nabla_{\mathbf{x}_t} \log p_{t}(\mathbf{x}_t)$:
$$
d \mathbf{x}_t=\underbrace{\left[\mathbf{f}(\mathbf x_t,t)-\frac{1}{2} g(t)^2 \mathbf s_{\boldsymbol{\theta}}(\mathbf{x}_t, t)\right]}_{=: \tilde{\mathbf{f}}_{\theta}(\mathbf{x}, t)} d t.
$$
With the instantaneous change of variables formula~\citep{Chen2018}, we can compute the log-likelihood of $p_0(\mathbf{x})$ using
$$
\log p_0(\mathbf{x}_0)=\log p_T(\mathbf{x}_T)+\int_0^T \nabla \cdot \tilde{\mathbf{f}}_{\theta}(\mathbf{x}_t, t) \mathrm{d} t
$$
where $\mathbf{x}(t)$ as a function of $t$ can be obtained by solving the probability flow ODE. Using $T=1$ and the definition of $\tilde{\mathbf{f}}_{\boldsymbol{\theta}}$ above, the log-likelihood is
$$
\log p_0(\mathbf{x}_0)=\log p_1(\mathbf{x}_1)+ \int_0^1 \left[\nabla \cdot\mathbf{f}(\mathbf x_t,t)-\frac{g(t)^2}{2} \nabla \cdot \mathbf s_{\boldsymbol{\theta}}(\mathbf{x}_t, t)\right] d t.
$$

\subsection{NLL Lower Bound}
The data-average log-likelihood at $t=0$ is
\begin{equation}
\mathbb{E}_{p_{\mathrm{data }}}\left[\log p_{\boldsymbol{\theta}}\left(\mathbf  x_0\right)\right]=\mathbb{E}_{p_1}\left[\log p_1\left(\mathbf x_1\right)\right]+ \int_0^1 \mathbb{E}_{p_t}\left[\nabla \cdot\mathbf{f}(\mathbf x_t,t)-\frac{g(t)^2}{2} \nabla \cdot \mathbf s_{\boldsymbol{\theta}}(\mathbf{x}_t, t)\right] dt.
\end{equation}
The first term is the entropy at $t=1$, $S_1=-\mathbb{E}_{p_1}\left[\log p_1\right]$. We re-express the divergence of the score term using $\mathbb{E}_{p_t}[\nabla \cdot \mathbf{s}_{\boldsymbol{\theta}}]=-\mathbb{E}_{p_t}\left[\mathbf{s}_{\boldsymbol{\theta}} \cdot \mathbf s_\text{true}\right]$ (Stein’s identity~\citep{Liu2016}), which gives:
$$
\mathrm{NLL}:=-\mathbb{E}_{p_{\mathrm{data }}}\left[\log p_{\boldsymbol{\theta}}\left(\mathbf x_0\right)\right]=S_1 + \int_0^1 \left[-\mathbb{E}_{p_t}\left[\nabla \cdot \mathbf{f}(\mathbf x_t,t)\right]-\frac{g(t)^2}{2} \mathbb{E}_{p_t}\left[\mathbf s_{\boldsymbol{\theta}} \cdot \mathbf s_\text{true}\right] \right] d t
$$

Using one of the polarization identities, $\mathbf s_{\boldsymbol{\theta}} \cdot \mathbf s^\text{true}=\frac{1}{2}\left(\left\|\mathbf s_{\boldsymbol{\theta}}\right\|^2+\left\|\mathbf s_\text{true}\right\|^2-\left\|\mathbf s_{\boldsymbol{\theta}}-\mathbf s_\text{true}\right\|^2\right)$, gives
\begin{align*}
    \mathrm{NLL}=S_1&-\int_0^1 \mathbb{E}_{p_t}\left[\nabla \cdot \mathbf{f}(\mathbf x_t,t)\right]dt-\frac{1}{2} \int_0^1 \frac{g(t)^2}{2} \mathbb{E}_{p_t}\left[||\mathbf s_{\boldsymbol{\theta}}||^2\right]dt\\&-\frac{1}{2} \int_0^1 \frac{g(t)^2}{2} \mathbb{E}_{p_t}\left[\mathbf ||\mathbf s_\text{true}||^2\right]dt+\frac{1}{2} \int_0^1 \frac{g(t)^2}{2} \mathbb{E}_{p_t}\left[||\mathbf s_{\boldsymbol{\theta}} - \mathbf s_\text{true}||^2\right] d t.
\end{align*}

We use $\int \frac{g(t)^2}{2}  \mathbb{E}\left\|\mathbf s_{\text {true }}\right\|^2=\left(S_1-S_0\right)-\int_0^1 \mathbb{E} [\nabla \cdot \mathbf f(\mathbf x_t,t)]dt$, giving
\begin{align*}
\mathrm{NLL}=\frac{S_0+S_1}{2}&-\frac{1}{2}\int_0^1 \mathbb{E}_{p_t}\left[\nabla \cdot \mathbf{f}(\mathbf x_t,t)\right]dt-\frac{1}{2} \int_0^1 \frac{g(t)^2}{2} \mathbb{E}_{p_t}\left[||\mathbf s_{\boldsymbol{\theta}}||^2\right]dt\\&+\frac{1}{2} \int_0^1 \frac{g(t)^2}{2} \mathbb{E}_{p_t}\left[||\mathbf s_{\boldsymbol{\theta}} - \mathbf s_\text{true}||^2\right] d t.
\end{align*}
where $S_0 = S(p_0) = S(p_\text{data})$. Using $\dot S_{\boldsymbol{\theta}}(t) = \mathbb{E}[\nabla_\mathbf{x}\cdot \mathbf f(\mathbf{x},t)]+\frac{g(t)^2}{2} \mathbb{E}\left[\|\mathbf s_{\boldsymbol{\theta}}(\mathbf x,t)\|^2\right]$ and the non-negativivity of the squared-difference term, we find that the negative log-likelihood obeys the lower bound
\begin{equation}
\boxed{\mathrm{NLL} \geq \frac{S_0+S_1}{2}-\frac{1}{2}\int_0^1 \dot S_{\boldsymbol{\theta}} (t)dt}
\end{equation}
and the bound is tight when $s_{\boldsymbol{\theta}} = s_\text{true}$ and $\text{NLL} =S_0$.

For the drift-less diffusion process, $\mathbf{f}(\mathbf x_t,t)=0$ and the system entropy rate is
$$
\dot{S}_{\boldsymbol{\theta}}(t) = \dot{S}_{\boldsymbol{\theta}}^i(t)=\frac{g(t)^2}{2} \mathbb{E}_{p_t}\left[\left\|\mathbf s_{\boldsymbol{\theta}}(\mathbf x, t)\right\|^2\right]
$$

so the lower bound is given in terms of entropies is
\begin{equation}
\boxed{\mathrm{NLL} \geq \frac{S_0+S_1}{2}-\frac{1}{2} \int_0^1 \dot{S}_{\boldsymbol{\theta}}^i(t) d t.}
\end{equation}

\subsection{Stein's Identity}\label{app:steins-identity}
In \citep{Liu2016}, Stein's identity states that for sufficiently regular $\phi$, we have
\begin{equation}
\mathbb{E}_{x \sim p}\left[\mathcal{A}_p \boldsymbol{\phi}(x)\right]=0, \quad \mathrm{ where } \quad \mathcal{A}_p \boldsymbol{\phi}(x)=\boldsymbol{\phi}(x) \nabla_x \log p(x)^{\top}+\nabla_x \boldsymbol{\phi}(x),
\end{equation}
where $\mathcal{A}_p$ is called the Stein operator, which acts on function $\phi$ and yields a zero mean function $\mathcal{A}_p \phi(x)$ under $x \sim p$. Expanding this identity coordinate-wise, it is exactly the statement:
$$
\mathbb{E}_p[\nabla \cdot \phi]=-\mathbb{E}_p[\phi \cdot s].
$$
With the true score $\mathbf s_\mathrm{true}=\nabla \log p(\mathbf x)$, we have:
$$
\mathbb{E}_p[\nabla \cdot \mathbf s_\mathrm{true}]=-\mathbb{E}_p\left[\|\mathbf s_\mathrm{true}\|^2\right].
$$
For an approximate score $\mathbf s_{\boldsymbol{\theta}}$ :
$$
\mathbb{E}_p\left[\nabla \cdot \mathbf s_{\boldsymbol{\theta}}\right]=-\mathbb{E}_p\left[\mathbf s_{\boldsymbol{\theta}} \cdot \mathbf s_\mathrm{true}\right]
$$
which equals $-\mathbb{E}_p\left[\left\|\mathbf s_{\boldsymbol{\theta}}\right\|^2\right]$ only if $\mathbf s_{\boldsymbol{\theta}}=\mathbf s_\mathrm{true}$.

\section{Maxwell’s Demon in Controlled-Forward Process}\label{app:maxwells-demon}
\cite{Song2021} use the notation of Haussman-Pardoux / Anderson~\citep{Haussmann1986,Anderson1982}
\begin{equation}
\mathrm{d} \mathbf{x}_t=\mathbf{f}(\mathbf{x},t)  \mathrm{d} t+g(t) \mathrm{d} \mathbf{w}_t \quad \mathrm{ (Forward) } 
\end{equation}
\begin{equation}
\mathrm{d} \mathbf{x}_t=[\mathbf{f}(\mathbf{x},t)  \mathrm{d} t-g(t)^2 \mathbf{s}_{\boldsymbol{\theta}}(\mathbf{x}, t)] \mathrm{d} t+g(t) \mathrm{d} \overline{\mathbf{w}}_t \quad \mathrm{ (Reverse) }
\end{equation}
where $\overline{\mathbf{w}}_t$ is the standard Wiener process when time is run backwards. \textbf{Note}, Eq. (8) is usually integrated from $T$ down to $0$, making $dt$ negative.

\subsection{Reverse Process}
We have chosen the forward SDE to be
$$
d \mathbf{x}_t=\sigma^t d \mathbf{w}_t, \quad t \in[0,1].
$$
To sample from our time-dependent score-based model $s_{\boldsymbol{\theta}}(\mathbf{x}, t)$, we first draw a sample from the prior distribution $p_1 \approx \mathbf{N}\left(\mathbf{x} ; \mathbf{0}, \frac{1}{2}\left(\sigma^2-1\right) \mathbf{I}\right)$, and then solve the reverse-time SDE with numerical methods. In particular, using our time-dependent score-based model, the reverse-time SDE can be approximated by
$$
d \mathbf{x}_t=-\sigma^{2 t} s_{\boldsymbol{\theta}}(\mathbf{x}, t) d t+\sigma^t d \overline{\mathbf{w}}_t
$$

Next, one can use numerical methods to solve for the reverse-time SDE, such as the Euler-Maruyama approach. It is based on a simple discretization to the SDE, replacing $d t$ with $\Delta t>0$ and $d \mathbf{w}$ with $\mathbf{z} \sim \mathcal{N}\left(\mathbf{0}, g^2(t) \Delta t \mathbf{I}\right)$. When applied to our reverse-time SDE, we can obtain the following iteration rule
$$
\mathbf{x}_{t-\Delta t}=\mathbf{x}_t+\sigma^{2 t} s_{\boldsymbol{\theta}}\left(\mathbf{x}_t, t\right) \Delta t+\sigma^t \sqrt{\Delta t} \mathbf{z}_t
$$
where $\mathbf{z}_t \sim \mathcal{N}(\mathbf{0}, \mathbf{I})$.

\subsection{Controlled-Forward Process}\label{app:controlled-forward}
Time always runs forward in the real world: one can achieve a physical realization of the generative process by defining a clock
$$
\tau:=T-t, \quad 0 \leq \tau \leq T,
$$
such that integrating forward in $\tau$ is the same as integrating backward in $t$. We plug in $t = T-\tau$, $dt = -d\tau$, and $d \bar{\mathbf w}_t=d \mathbf w _\tau$ to re-parameterize reverse Eq.~(8) as a controlled-forward process
\begin{equation}
\mathrm{d} \bar{\mathbf{x}}_\tau=[-\mathbf f(\bar{\mathbf{x}}_\tau,T-\tau)  +g(T-\tau)^2 \mathbf{s}_{\boldsymbol{\theta}}(\bar{\mathbf{x}}_\tau, T-\tau)] \mathrm{d} \tau+g(T-\tau) \mathrm{d} \mathbf{w}_\tau 
\end{equation}
where $\bar {\mathbf x}_\tau := \mathbf x_{t}$.

\subsection{Entropy Rates of the Controlled-Forward Process}
For the controlled-forward formulation, the drift becomes $\tilde{\mathbf{f}}(\bar{\mathbf x}_\tau,\tau) = g(\tau)^2 \mathbf s_{\boldsymbol{\theta}}(\bar{\mathbf x}_\tau,\tau)$. 
Substituting this into the general entropy rate expressions derived in Appendix~\ref{app:entropy-rates}
yields the simplified relations:
\[
\dot S^i_{\boldsymbol{\theta}}(\tau) = \frac{g(\tau)^2}{2}\, \mathbb{E}\!\left[\|\mathbf s_{\boldsymbol{\theta}}(\bar{\mathbf x}_\tau,\tau)\|^2\right], \quad
\dot S^e_{\boldsymbol{\theta}}(\tau) = -2 \dot S^i_{\boldsymbol{\theta}}(\tau), \quad
\dot S_{\boldsymbol{\theta}}(\tau) = -\dot S^i_{\boldsymbol{\theta}}(\tau).
\]
Thus, in the controlled-forward process the system entropy rate is exactly the negative of the intrinsic entropy production rate.

\section{General Continuous-Time Diffusion Processes}
\label{app:vp}
\citet{Song2021} showed that score-based generative models can be formulated in terms of a general Itô SDE of the form
$$
d \mathbf x_t=\mathbf f\left(\mathbf x_t, t\right) d t+g(t) d \mathbf w_t
$$
where $\mathbf f\left(\mathbf x_t, t\right)$ is the drift, $g(t)$ the diffusion coefficient, and $\mathbf w_t$ a standard Wiener process. Two canonical instantiations of this framework correspond to the variance exploding (VE) and variance preserving (VP) processes.

\subsection{Variance Exploding (VE) SDE}
The VE process is defined by
$$
d \mathbf x_t=\sqrt{\frac{d}{d t} \sigma^2(t)} d \mathbf w_t
$$
with $\sigma^2(t)$ a non-decreasing variance schedule. Here the drift vanishes, $\mathbf f\left(\mathbf x_t, t\right)=0$, while the diffusion coefficient is chosen so that the marginal variance of $\mathbf x_t$ increases monotonically in $t$. As $t \rightarrow T$, the variance diverges (hence "exploding"), and the distribution approaches a Gaussian prior. This setting is natural when starting from bounded data distributions, since the forward process progressively washes out structure by injecting unbounded noise

\subsection{Variance Preserving (VP) SDE}
In contrast, the VP process includes both drift and diffusion terms:
$$
d \mathbf x_t=-\frac{1}{2} \beta(t) \mathbf x_t d t+\sqrt{\beta(t)} d \mathbf w_t
$$
where $\beta(t)$ is a positive noise-rate schedule. The drift pulls $\mathbf x_t$ toward the origin at a rate proportional to $\beta(t)$, while the diffusion injects noise of matching strength. This balance ensures that the overall variance of the process remains bounded (and can be normalized to unity) for all $t$. Thus, the forward diffusion maps data smoothly into an isotropic Gaussian prior without variance blow-up.

Both SDEs fit seamlessly into the score-based generative modeling framework. In each case, the reverse-time dynamics introduce an additional score-dependent drift term,
$$
d \mathbf x_\tau=\left[-\mathbf f\left(\mathbf x_\tau, T-\tau\right)+g^2(T-\tau) \nabla_{\mathbf x_\tau} \log p_\tau\left(\mathbf x_\tau\right)\right] d \tau+g(T-\tau) d \mathbf w_\tau
$$
where the score $\nabla_{\mathbf x_\tau} \log p_\tau\left(\mathbf x_\tau\right)$ is approximated by a neural network. The VE and VP choices thus represent two distinct, yet complementary, continuous-time noise injection schemes, both of which reduce to the driftless case when $f=0$ and variance is allowed to grow freely. They provide the practical foundation for most modern diffusion models, differing primarily in how variance is managed over time and, correspondingly, in their tradeoffs between sample quality and likelihood.

In addition to the variance exploding (VE) process considered in the main text, we evaluate the variance preserving (VP) process used in denoising diffusion probabilistic models (DDPMs). The VP process is governed by the forward SDE
\[
d \mathbf{x}_t = -\tfrac{1}{2}\beta(t)\,\mathbf{x}_t\,dt + \sqrt{\beta(t)}\,d\mathbf{w}_t,
\]
where $\beta(t)$ is the variance schedule and $w_t$ is standard Brownian motion. This process interpolates between the data distribution at $t=0$ and an isotropic Gaussian prior at $t=1$ while preserving variance at each time step. The reverse process is parameterized by the learned score network, and the associated entropy-production integrals are estimated analogously to the VE case.

\subsection{Constructing the VP Schedule from $\sigma$}
To specify $\beta(t)$, we set a desired terminal noise scale $\sigma$, which encodes how much Gaussian noise is injected by the end of the forward process.

\subsubsection{Integrated noise budget}
The mean-scaling factor of the VP process is
$$
\alpha(t)=\exp \left(-\frac{1}{2} \int_0^t \beta(u) d u\right)
$$
so at terminal time $t=1$,
$$
\alpha(1)^2=\exp (-B), \quad B:=\int_0^1 \beta(u) d u
$$
The variance contributed by the noise term is
$$
\sigma(1)^2=1-\alpha(1)^2 .
$$
Requiring $\sigma(1)^2=\sigma^2/1+\sigma^2$ yields the condition
$$
B=\log \left(1+\sigma^2\right)
$$
Thus the entire VP schedule is determined by the integrated noise budget $B$.

\subsubsection{Linear schedule construction}
A common choice is to make $\beta(t)$ linear in $t$:
$$
\beta(t)=\beta_{\min }+t\left(\beta_{\max }-\beta_{\min }\right)
$$
The constants $\beta_{\text {min }}$ and $\beta_{\text {max }}$ are set so that the integral matches the budget:
$$
\int_0^1 \beta(t) d t=\frac{1}{2}\left(\beta_{\min }+\beta_{\max }\right)=B
$$
Introducing a ratio parameter $r \in(0,1)$, we define
$$
\beta_{\min }=r B, \quad \beta_{\max }=(2-r) B,
$$
which ensures the correct average while allowing flexibility in the temporal profile of noise injection. Smaller $r$ front-loads noise near $t=1$, while larger $r$ distributes noise more evenly across time.

In the VP formulation, $B$ is the logarithmic noise budget: it quantifies the total exponential damping of the signal. In the VE formulation, the corresponding budget is the variance scale $\sigma^2$. The two are linked by $\sigma^2=e^B-1$. Hence, the VP schedule can be constructed from a single intuitive parameter $\sigma$, which specifies the effective strength of the forward noise process, while $B$ serves as its natural exponential coordinate.

\section{Standard Gaussian Data}
\label{app:gaussian_vp}
For validation we include experiments where the data distribution $p_{\text{data}}$ is a standard Gaussian. In this case, the score function is exactly linear:
\[
\nabla_\mathbf{x} \log p_{\text{data}}(\mathbf{x}) = -\mathbf{x},
\]
which can be fit by a single-layer neural network with linear weights. This setting provides a ground-truth baseline where the score is known analytically, allowing us to verify the tightness of the lower bound and the accuracy of our numerical estimators.

Consider the case in which the data is normally distributed $\mathbf x_0 \sim \mathcal{N}(\boldsymbol \mu, \boldsymbol \Sigma)$ and a drift-less forward process with variance increment $v(t)=\int_0^t g(u)^2 d u$, with $g^2(t)=\sigma^{2 t}$ and $v(t)=\frac{\sigma^{2 t}-1}{2 \ln \sigma}$. Then $\mathbf x_t \sim \mathcal{N}(\boldsymbol \mu, \boldsymbol \Sigma+v(t) \boldsymbol I)$ and
$$
\mathbf s_{\text {true }}(\mathbf x, t)=\nabla_{\mathbf x} \log p_t(\mathbf x)=-(\boldsymbol \Sigma+v(t) \mathbf I)^{-1}(\mathbf x-\boldsymbol \mu),
$$
which is exactly linear in $\mathbf x$ for every $t$. A tiny network (even a single linear layer conditioned on $t$) can represent this perfectly, so training can drive $\mathbf s_{\boldsymbol \theta} \rightarrow \mathbf s_{\text {true }}$.

\textit{Proof}.  Let
$$
p_t(\mathbf x)=\mathcal{N}(\mathbf x ; \boldsymbol \mu, \mathbf C(t)), \quad \mathbf C(t)=\boldsymbol \Sigma+v(t) \mathbf I_d,
$$
so $\mathbf C(t)$ is symmetric positive-definite. The multivariate Gaussian pdf is
$$
p_t(\mathbf x)=\frac{1}{(2 \pi)^{d / 2} \operatorname{det}(\mathbf C)^{1 / 2}} \exp \left(-\frac{1}{2}(\mathbf x-\boldsymbol \mu)^{\top} \mathbf C^{-1}(\mathbf x-\boldsymbol \mu)\right)
$$
Take logs:
$$
\log p_t(\mathbf x)=-\frac{1}{2}(\mathbf x-\boldsymbol \mu)^{\top} \mathbf C^{-1}(\mathbf x-\boldsymbol \mu)-\frac{1}{2} \log \operatorname{det}(2 \pi \mathbf C) .
$$
Only the quadratic term depends on $\mathbf x$. With $h=(x-\mu)_j A_{j k}(x-\mu)_k$,
$$
\frac{\partial h}{\partial x_i}=A_{i j}(x-\mu)_j+A_{j i}(x-\mu)_j=\left[\left(A+A^{\top}\right)(x-\mu)\right]_i
$$
and we have
$$
\nabla_{\mathbf x}\left[(\mathbf x-\boldsymbol \mu)^{\top} \mathbf A(\mathbf x-\boldsymbol \mu)\right]=\left(\mathbf A+\mathbf A^{\top}\right)(\mathbf x-\boldsymbol \mu)=2 \mathbf A(\mathbf x-\boldsymbol \mu) \quad\left(\mathbf A=\mathbf A^{\top}\right),
$$
with $\mathbf A=\mathbf C^{-1}$, we get
$$
\nabla_{\mathbf x} \log p_t(\mathbf x)=-\frac{1}{2} \cdot 2 \mathbf C^{-1}(\mathbf x-\boldsymbol \mu)=-\mathbf C^{-1}(\mathbf x-\boldsymbol \mu).
$$
Therefore the true score is
$$
\mathbf s_{\text {true }}(\mathbf x, t)=\nabla_{\mathbf x} \log p_t(\mathbf x)=-(\boldsymbol \Sigma+v(t) \mathbf I)^{-1}(\mathbf x-\boldsymbol \mu)
$$
exactly as claimed.

\section{Exact Score of the Uniform + Normal Distributions}
Pixels are independent under Uniform[0,1] and the Gaussian noise factorizes across coordinates, so we consider the 1D case and adopt scalar notation throughout. For the drift-less diffusion,
$$
d x_t=g(t) d W_t, \quad v(t)=\int_0^t g(u)^2 d u, \quad s=\sqrt{v(t)}
$$

Conditioned on $x_0$, we have
$$
x_t \mid x_0 \sim \mathcal{N}\left(x_0, s^2\right)
$$

If the data is Uniform on $[0,1]$ (density $p_0(u)=\mathbf{1}_{[0,1]}(u)$), the marginal at time $t$ is the convolution
$$
p_t(x)=\int_0^1 \phi_s(x-u) d u
$$
where 
$$
\phi_s(z)=\frac{1}{\sqrt{2 \pi} s} \exp \left(-\frac{z^2}{2 s^2}\right)
$$ 
is the $\mathcal{N}\left(0, s^2\right)$ pdf. With a change of variable $z=(x-u) / s$ and $d u=-s d z$, we have
$$
p_t(x)=\int_{(x-1) / s}^{x / s} \phi(z) d z=\Phi\left(\frac{x}{s}\right)-\Phi\left(\frac{x-1}{s}\right)
$$
with $\phi$ and $\Phi$ the standard normal pdf/cdf.
Differentiate w.r.t. $x$ :
$$
\partial_x p_t(x)=\frac{1}{s}\left[\phi\left(\frac{x}{s}\right)-\phi\left(\frac{x-1}{s}\right)\right] . 
$$
The score is the gradient of the log-density,
$$
s(x, t)=\partial_x \log p_t(x)=\frac{\partial_x p_t(x)}{p_t(x)}=\frac{\frac{1}{s}\left[\phi\left(\frac{x}{s}\right)-\phi\left(\frac{x-1}{s}\right)\right]}{\Phi\left(\frac{x}{s}\right)-\Phi\left(\frac{x-1}{s}\right)}. 
$$

\section{Numerical Estimates of Exact NLL Terms}
As summarized in Table~\ref{table:error-sources}, the decomposition includes both exact and estimated terms, with dominant error sources arising from finite-batch sampling, quadrature, and model fit.

\subsection{Equilibrium entropy $S_1$}
At $t=1$, the forward drift-less diffusion process has covariance $v(1) \mathbf I$ with
$$
v(1)=\frac{\sigma^2-1}{2 \ln \sigma} .
$$
Hence the equilibrium entropy in nats for a $d$-dimensional Gaussian is
$$
S_{1, \mathrm{nats}}=\frac{d}{2} \ln (2 \pi e v(1))=\frac{d}{2} \ln \left(2 \pi e \frac{\sigma^2-1}{2 \ln \sigma}\right)
$$
and in bits-per-dimension (bpd),
$$
S_{1}=\frac{S_{1, \mathrm{nats}}}{d \ln 2}
$$
This term is computed analytically. We compute the exact closed form and convert it to bpd.

\subsection{Dataset entropy $S_0$}
These constants are independent of the diffusion schedule (VE vs VP) and enter directly into the NLL lower bound formulas. For the standard Gaussian datasets considered, the data entropy is fixed by the closed-form expression
$$
S_0=\frac{1}{2} d \log (2 \pi e)
$$
corresponding to the entropy of a $d$-dimensional standard normal. 

For the Uniform $[0,1]^d$ datasets, the entropy vanishes, $S_0=0$, since the density is constant on its support. Let $X\sim \mathrm{Unif}([0,1]^d)$, so $p(x)=1$ for $x\in[0,1]^d$ and $p(x)=0$ otherwise.
The differential entropy is
\[
S_0 \;=\; -\int_{\mathbb{R}^d} p(x)\,\log p(x)\,dx
\;=\; -\int_{[0,1]^d} 1\cdot \log 1 \; dx
\;=\; 0.
\]
By factorization across coordinates,
$S_0 = \sum_{i=1}^d h(X_i)$ with $X_i\sim\mathrm{Unif}([0,1])$ and
$h(X_i) = -\!\int_0^1 1\cdot \log 1\, dx_i = 0$, hence $S_0=0$.

\subsection{Squared-norm of the model score, $I_{\boldsymbol{\theta}}$}
We estimate
$$
I_{\boldsymbol{\theta}}=\frac{1}{2} \int_0^1 g(t)^2 \mathbb{E}_{\mathbf x_t \sim p_t}\left[\left\|\mathbf s_{\boldsymbol{\theta}}\left(\mathbf x_t, t\right)\right\|^2\right] d t
$$

Estimator (per time grid $t_k$, batch size $B$):
\begin{enumerate}
    \item Draw $\mathbf x_0 \sim p_{\mathrm{data }}, \mathbf z \sim \mathcal{N}(0, \mathbf I)$.
    \item Form $\mathbf x_t=\mathbf x_0+\sqrt{v\left(t_k\right)} \mathbf z$.
    \item Evaluate the model score $\mathbf s_{\boldsymbol{\theta}}\left(\mathbf x_t, t_k\right)$.
    \item Compute the batch mean $\widehat{E}_k=\frac{1}{B} \sum_{i=1}^B\left\|\mathbf s_{\boldsymbol{\theta}}\left(\mathbf x_t^{(i)}, t_k\right)\right\|^2$.
\end{enumerate}
Finally, we integrate over $t$ with the trapezoid rule:
$$
\hat{I}_{\boldsymbol{\theta}}=\frac{1}{2} \sum_k w_k \widehat{E}_k, \quad w_k=g\left(t_k\right)^2 \Delta t_k
$$
and convert to bpd by dividing by $d \ln 2$.

\subsection{Squared-difference term, $I_{\mathrm{diff }}$}
$$
I_{\mathrm{diff }}=\frac{1}{2} \int_0^1 g(t)^2 \mathbb{E}_{\mathbf x_t \sim p_t}\left[\left\|\mathbf s_{\boldsymbol{\theta}}\left(\mathbf x_t, t\right)-\mathbf s_{\mathrm{true }}\left(\mathbf x_t, t\right)\right\|^2\right] d t \geq 0
$$
We estimate it in two ways: directly by computing $\left\|\mathbf s_{\boldsymbol{\theta}}-\mathbf s_{\mathrm{true }}\right\|^2$ per sample and average, and using the polarization identity as a sanity check:
$$
\|a-b\|^2=\|a\|^2+\|b\|^2-2\langle a, b\rangle
$$
to obtain the the squared difference term from separate estimates of $\left\|\mathbf s_{\boldsymbol{\theta}}\right\|^2,\left\|\mathbf s_{\mathrm{true }}\right\|^2$, and $\left\langle \mathbf s_{\boldsymbol{\theta}}, \mathbf s_{\mathrm{true }}\right\rangle$. We use the agreement between the two as a useful consistency diagnostic.

\subsection{Error sources}
\subsubsection{Finite-batch Monte-Carlo error}
For a fixed $t_k$, the batch mean $\widehat{E}_k$ is an unbiased estimator of $\mathbb{E}[\cdot]$ with variance $\operatorname{Var}\left[\widehat{E}_k\right]=\operatorname{Var}[\cdot] / B$.
Propagating through the trapezoid rule gives an approximate variance
$$
\operatorname{Var}[\hat{I}] \approx \frac{1}{4} \sum_k w_k^2 \frac{\operatorname{Var}_{p_{t_k}}\left[U\left(x_{t_k}, t_k\right)\right]}{B}, \quad U \in\left\{\left\|\mathbf s_{\boldsymbol{\theta}}\right\|^2,\left\|\mathbf s_{\mathrm{true }}\right\|^2,\left\|\mathbf s_{\boldsymbol{\theta}}-\mathbf s_{\mathrm{true }}\right\|^2\right\} .
$$

\subsubsection{Time-integration (quadrature) error}
With a smooth integrand, the trapezoid rule has $O\left(\Delta t^2\right)$ bias. Our implementation guards common pitfalls: it clips $t \in\left[10^{-4}, 1-10^{-4}\right]$ to avoid extreme-variance endpoints and integrates with NumPy's trapezoidal function.

\subsubsection{Goodness-of-fit (modeling) error}
Only terms that involve $s_{\boldsymbol{\theta}}$ suffer approximation error:
\begin{itemize}
    \item $I_{\boldsymbol{\theta}}$ equals the target $I_{\mathrm{true }}$ iff $\mathbf s_{\boldsymbol{\theta}} \equiv \mathbf s_{\mathrm{true }}$. The gap is not simply $I_{\mathrm{diff }}$ due to the cross-term $\int g(t)^2 \mathbb{E}\left\langle \mathbf s_{\boldsymbol{\theta}}-\mathbf s_{\mathrm{true }}, \mathbf s_{\mathrm{true }}\right\rangle d t$. In practice we monitor $I_{\mathrm{diff }}$ (nonnegative, zero at optimum) and cosine similarity of $\mathbf s_{\boldsymbol{\theta}}$ vs. $s_{\mathrm{true }}$ as diagnostics (our code logs min/mean/max cosine).
    \item $I_{\mathrm{diff }}$ itself is zero iff the model is perfect; otherwise it is positive and captures a portion of the model-fit error. When $s_{\mathrm{true }}$ is unavailable (e.g., MNIST), any proxy introduces additional modeling bias on top of Monte-Carlo and quadrature error.
\end{itemize}

\subsubsection{Numerical error: floating-point and conditioning}
Computing $\mathbf s_{\mathrm{true }}$ for the variance-expanded uniform uses $\Phi\left(z_L\right)-\Phi\left(z_R\right)$; we clamp the denominator and evaluate in float64 to avoid catastrophic cancellation when $t$ is small/large. The network outputs are float32; we upcast to float64 before inner products, which prevents accumulation error in norms and inner products.

\begin{table}[h]
\centering
\begin{tabular}{lp{0.5\textwidth}l}
\toprule
\textbf{Quantity} & \textbf{Approach} & \textbf{Error type(s)} \\
\midrule
$H_{1,\mathrm{bpd}}$ 
  & Closed-form Gaussian entropy at $t=1$ 
  & Exact \\[6pt]

$I_{\theta,\mathrm{bpd}}$ 
  & Monte Carlo over $(\mathbf x_0,\mathbf z)$ + quadrature of $\|\mathbf s_{\boldsymbol{\theta}}\|^2$ 
  & Finite-batch; quadrature; model fit \\[6pt]

$I_{\mathrm{diff},\mathrm{bpd}}$ 
  & Same, using $\|\mathbf s_{\boldsymbol{\theta}} - \mathbf s_{\mathrm{true}}\|^2$ (and polarization check) 
  & Finite-batch; quadrature; model fit \\[6pt]
\bottomrule
\end{tabular}
\caption{Summary of which terms in the NLL decomposition are exact vs. estimated, and their dominant sources of error.}
\end{table}\label{table:error-sources}

\end{document}